# Inferring Parameters and Structure of Latent Variable Models by Variational Bayes

**Hagai Attias**
hagai@gatsby.ucl.ac.uk
Gatsby Unit, University College London
17 Queen Square
London WC1N 3AR, U.K.

## Abstract

Current methods for learning graphical models with latent variables and a fixed structure estimate optimal values for the model parameters. Whereas this approach usually produces overfitting and suboptimal generalization performance, carrying out the Bayesian program of computing the full posterior distributions over the parameters remains a difficult problem. Moreover, learning the structure of models with latent variables, for which the Bayesian approach is crucial, is yet a harder problem. In this paper I present the Variational Bayes framework, which provides a solution to these problems. This approach approximates full posterior distributions over model parameters and structures, as well as latent variables, in an analytical manner without resorting to sampling methods. Unlike in the Laplace approximation, these posteriors are generally non-Gaussian and no Hessian needs to be computed. The resulting algorithm generalizes the standard Expectation Maximization algorithm, and its convergence is guaranteed. I demonstrate that this algorithm can be applied to a large class of models in several domains, including unsupervised clustering and blind source separation.

## 1 Introduction

This paper focuses on learning graphical models from data. A standard method to learn a model is maximum likelihood (ML). This method estimates optimal values for the model parameters within a fixed graph structure from a given dataset. There are three main problems with ML learning. First, it produces a model that overfits the data and subsequently have suboptimal generalization performance. Second, it cannot be used to learn the structure of the graph, since more complicated graphs assign a higher likelihood to the data. Third, it is computationally tractable only for a small class of models.

The Bayesian framework (Mackay 1992a, 1992b; Cooper and Herskovits 1992; Heckerman et al. 1995) provides, in principle, a solution to the first two problems. In this framework one considers an ensemble of models, characterized by a probability distribution over all possible parameter values and structures. Rather than learning a single model from a given dataset, one computes the distribution over the ensemble of models given these data. Model uncertainty is thus taken into account, leading to enhanced generalization performance. In addition, complex models are effectively penalized by being assigned a lower posterior probability, hence optimal structures can be identified. In models that contain hidden variables, the posterior one computes is the joint distribution over models and hidden variables given the data.

Unfortunately, computations in the Bayesian framework can seldom be performed exactly, due to the need to integrate over models. Approximations therefore must be made (see, e.g., Cheeseman and Stutz 1995; Chickering and Heckerman 1997; Friedman 1998), the major schemes being Markov chain Monte Carlo methods and Laplace approximation. The former attempts to achieve exact results but typically requires vast computational resources. The latter has lower complexity of $\mathcal{O}(m^2 N)$, where $m$ is the number of parameters and $N$ the dataset (sample) size, but is a good approximation only in the limit $N/m \to \infty$; in particular, is assumes that all posterior distributions are Normal (but see the discussion in Mackay 1998a). Naturally, the situation becomes worse when hidden variables exist.

In this paper I present the Variational Bayes framework for computations in graphical models. This framework facilitates analytical calculations of poste-rior distributions over the hidden variables, parameters and structures. It draws together variational ideas from intractable hidden variables models (Saul, Jaakkola and Jordan 1996; Ghahramani and Jordan 1997) and from Bayesian inference (Waterhouse, Mackay and Robinson 1996; Jaakkola & Jordan 1997; Mackay 1998), which, in turn, draw on the work of Neal and Hinton (1998). The posteriors are obtained via an iterative EM-like algorithm whose convergence is guaranteed. Focusing on the parameter posterior, its resulting approximation is more efficient than the Laplace as the Hessian needs not be computed, and produces non-trivial posteriors for any sample size. In addition, the BIC/MDL model selection criteria are obtained from VB in the large sample limit. The VB framework is developed in section 2, and is applied to mixture models in section 3 and to the blind source separation problem in section 4. Learning structure of complex models is discussed in section 5.

**Notation.** We shall use the Dirichlet, Normal, and Wishart distributions $\mathcal{D}$, $\mathcal{N}$, $\mathcal{W}$ in the following parametrization:

$$\mathcal{D}(\{\pi_s\}; \{\lambda_s\}) \propto \prod_{s=1}^{m} \pi_s^{\lambda_s - 1},$$
$$\mathcal{N}(\mathbf{x}; \boldsymbol{\mu}, \boldsymbol{\Sigma}) \propto e^{-(\mathbf{x}-\boldsymbol{\mu})^T \boldsymbol{\Sigma} (\mathbf{x}-\boldsymbol{\mu})/2},$$
$$\mathcal{W}(\boldsymbol{\Gamma}; a, \mathbf{B}) \propto |\boldsymbol{\Gamma}|^{a-1} e^{-\text{Tr} \mathbf{B} \boldsymbol{\Gamma}}. \quad (1)$$

Note that $\boldsymbol{\Sigma}$ is the inverse covariance (a.k.a. precision) matrix of $\mathcal{N}$. We also use the Normal-Wishart distribution

$$\mathcal{NW}(\mathbf{x}, \boldsymbol{\Gamma}; a, \mathbf{B}, \boldsymbol{\mu}, \beta) = \mathcal{W}(\boldsymbol{\Gamma}; a, \mathbf{B}) \mathcal{N}(\mathbf{x}; \boldsymbol{\mu}, \beta \boldsymbol{\Gamma}). \quad (2)$$

## 2 The Variational Bayes Framework for Graphical Models

### 2.1 Definitions

We restrict our attention in this paper to directed acyclic graphs (a.k.a. Bayesian networks). Let $M$ denote a set of model structures. The variables in a structure $m \in M$ are divided into two sets: visible (data) variables $v_i \in V$ and hidden (latent) variables $h_i \in H$. Each variable is a vector of some dimension, whose coordinates may assume either discrete or continuous values. A structure $m$ specifies (a) the visible set $V$, which is the same for all models; (b) the hidden set $H$; (c) the dependencies (i.e., directed edges) between the variables; and (d) the parametrized form of the probabilistic dependencies $p(u_i \mid \mathbf{pa}_i, \theta_i, m)$, where $u_i \in V \cup H$, $\mathbf{pa}_i$ is the set of parents of $u_i$, and $\theta_i$ denotes the parameter set of this conditional distribution. Hence, different structures may have different numbers of hidden variables, and a given hidden variable may have a different dimensionality or assume a different set of values in different structures. The reason we consider vector variables rather than the customary scalars is that we shall occasionally be using real-valued distributions $p(u_i \mid \mathbf{pa}_i, \theta_i, m)$ that allow correlations between the coordinates of $u_i$, which are therefore not conditionally independent; of course, this could have also been achieved using a slightly more complicated graph.

Denoting the complete set of parameters by $\Theta = \cup_i \{\theta_i\}$, the relevant joint distribution is

$$p(V, H, \Theta, m) = \prod_i p(u_i \mid \mathbf{pa}_i, \theta_i, m) p(\Theta \mid m) p(m) \quad (3)$$

where $p(\Theta \mid m)$ is the prior distribution on the parameters of structure $m$ and $p(m)$ is the prior over our set of structures. As a final note on terminology, the term *model* will refer to a pair $(\Theta, m)$ of a specific structure and specific parameter values.

We are interested in the *ensemble likelihood* $p(Y)$. This is the likelihood $\prod_n p(V = \mathbf{y}_n \mid \Theta, m)$ assigned to the dataset $Y = \{\mathbf{y}_n, n = 1 : N\}$ by model $(\Theta, m)$, averaged over the ensemble of models described by $p(\Theta, m) = p(\Theta \mid m) p(m)$. This quantity is also known as *marginal likelihood* or *evidence*. Note that its calculation requires averaging over all configurations of the hidden units within each model. The ensemble likelihood is generally computationally intractable, as it requires (a) integrating the joint (3) over the parameters, which typically cannot be performed analytically; (b) summation over all possible values of the hidden variables. For discrete variables, the number of terms in these sums is exponential in the number of nodes, whereas for real-valued variables these sums may turn into analytically intractable integrals; (c) summation over all possible structures, whose number grows exponentially with the maximum numbers allowed for nodes and edges. In the following we address these issues within a variational framework.

### 2.2 Ensemble Likelihood and Occam Factor

The Variational Bayes framework is formulated as follows. Starting from the ensemble liklehood, we use the Neal-Hinton representation (Neal and Hinton 1998) to place a lower bound on it:

$$\mathcal{L} = \log p(Y) \geq \mathcal{F} \equiv \sum_{m \in M} \int d\Theta \sum_H$$
$$q(H, \Theta, m \mid Y) \log \frac{p(Y, H, \Theta, m)}{q(H, \Theta, m \mid Y)}, \quad (4)$$

where the sum over $H$ ranges over all possible values of all hidden variables and implies integration for continuous variables. The inequality (4) holds for an arbitrary conditional distribution $q$. The optimal $q$ is obtained by setting the functional derivative of the right hand side with respect to $q$ to zero; the resulting equation is solved only by the true posterior, $q = p(H, \Theta, m \mid Y)$, obtained from (3) using Bayes' rule. It is easy to show that in this case the inequality (4) becomes an equality. However, the computation of the true posterior is intractable and approximations must be made. Our approach restricts the space of allowed $q$ to distributions where the parameters are conditionally independent of the hidden variables given a structure, i.e., have the form

$$q(H, \Theta, m \mid Y) = q(H \mid m, Y) q(\Theta \mid m, Y) q(m \mid Y) . \quad (5)$$

This posterior will generally differ from the true one and is termed the *variational posterior*. It will be optimized to produce the best approximation to the true posterior. Hence, we get a lower bound $\mathcal{F}$ on the ensemble likelihood, which splits into two terms:

$$\begin{aligned}
\mathcal{F} &= \mathcal{F}_{\Theta,m} - \mathcal{D}_{\Theta,m} , \\
\mathcal{F}_{\Theta,m} &= \langle \sum_H q(H \mid m) \log \frac{p(Y, H \mid \Theta, m)}{q(H \mid m)} \rangle_{\Theta,m} , \\
\mathcal{D}_{\Theta,m} &= \mathrm{KL}[q(\Theta, m) \parallel p(\Theta, m)] , \quad (6)
\end{aligned}$$

where the average $\langle \cdot \rangle_{\Theta,m}$ in the first term is computed with respect to the model posterior $q(\Theta, m)$, and the second term is the Kullback-Leibler (KL) distance between $q(\Theta, m)$ and $p(\Theta, m)$, i.e., $\mathcal{D}_{\Theta,m} = \langle \log q(\Theta,m)/p(\Theta,m) \rangle_{\Theta,m}$. The dependence of the variational posteriors on the data $Y$ is henceforth omitted. As we shall see, the first term corresponds to the likelihood term, whereas the second term is the Occam factor which penalizes for over complex models. Thus, our score function $\mathcal{F}$ may be interpreted as a penalized likelihood, where *the penalty is the KL distance between the posterior and prior distributions over the ensemble of models*.

### 2.3 Large Sample Limit

To gain some insight into $\mathcal{F}$, consider the large sample limits $N \to \infty$. In this case, the model posterior is strongly peaked about its mean, the maximum likelihood (ML) model $(\Theta_0, m_0)$, with a covariance that typically decreases as $1/N$. To compute the first term of $\mathcal{F}$ in this limit, only the ML model $(\Theta_0, m_0)$ needs to be included; the relative correction will be $\mathcal{O}(1/N)$. To compute the second term, we approximate $q(\Theta, m_0)$ by a multivariable Gaussian distribution with the same mean and covariance. We thus obtain

$$\begin{aligned}
\mathcal{F}(N \to \infty) &= \mathcal{F}_0 - \mathcal{D}_0 , \\
\mathcal{F}_0 &= \sum_H q(H) \log \frac{p(Y, H \mid \Theta_0)}{q(H)} ,
\end{aligned}$$

$$\mathcal{D}_0 = \frac{|\Theta_0|}{2} \log N - \log p(\Theta_0) , \quad (7)$$

where $|\Theta_0|$ is the number of parameters in the ML model, and the $m_0$ dependence is omitted. Let us first focus on maximizing $\mathcal{F}_0$ alone. As shown in (Neal and Hinton 1998), this is a generalized representation of the ML problem where we seek a single parameter value $\Theta_0$. The ordinary EM algorithm is obtained by maximizing $\mathcal{F}_0$ w.r.t. $q(H)$ and $\Theta_0$ alternately: In the E-step of the $r$th iteration we set $\delta \mathcal{F}_0 / \delta q(H) = 0$, which gives $q(H) = p(H \mid Y, \Theta_0^{(r-1)})$; in the M-step we fix $q$ and solve $\partial \mathcal{F}_0 / \partial \Theta_0 = 0$ to obtain $\Theta_0^{(r)}$. The variational EM algorithm (Saul et al. 1996) was introduced for cases where the computation of the exact posterior $p(H \mid Y)$ is intractable. Instead, a form $q(H)$ which allows performing the calculation is used; it has its own set of parameters, and in the E-step these parameters are optimized to minimize the KL distance between $q(H)$ and the true posterior.

Second, the penalty $\mathcal{D}_0$ reduces in this limit to a term that is linear in the number of the ML model parameters, plus a simple regularizer $-\log p(\Theta_0)$. Finally, we point out that the Bayesian information criterion (BIC) (Schwartz 1978) and the minimum description length criterion (MDL) (Rissanen 1987) both emerge as a special case of our large sample expression (7), corresponding to using flat prior $p(\Theta)$ and exact (rather than variational) posterior $q(H)$.

### 2.4 Optimal Posteriors and Relation to EM

To find the optimal variational posterior over the parameters for a given structure $m$, we set $\delta \mathcal{F}/\delta q(\Theta \mid m) = 0$ in (6) and obtain

$$\begin{aligned}
\log q(\Theta \mid m) &= \langle \log p(Y, H \mid \Theta, m) \rangle_{H \mid m} \\
&\quad + \log p(\Theta \mid m) - \log z_m , \quad (8)
\end{aligned}$$

where the average $\langle \cdot \rangle_{H \mid m}$ in the first term is computed w.r.t. the hidden variable posterior $q(H \mid m)$, and $z_m$ is a normalization constant. In spite of the apparent complexity of (8), the resulting posterior is typically quite simple. First, averaging over the hidden variables using the variational posterior $q(H \mid m)$ is obtained in a closed form; this is a key property of the variational approach (see below).

Second, if we use directed graphs where each node has its own parameters, and if we use a parameter prior that factorizes over the nodes, then the parameter posterior factorizes as well. To see this, recall the joint distribution over the nodes (3), and assume $p(\Theta \mid m) = \prod_i p(\theta_i \mid m)$. From (8) we then have

$$\log q(\Theta \mid m) = \sum_i \langle \log p(u_i \mid \mathbf{pa}_i, \theta_i, m) \rangle_{H \mid m}$$

$$+ \sum_i \log p(\theta_i \mid m) - \log z_m$$
$$\equiv \sum_i \log q(\theta_i \mid m) \,, \qquad (9)$$

proving that, given a particular graph structure, the posteriors over the parameters of different nodes are mutually independent.

Third, the functional form of the parameter posterior is determined by the distributions that define our model, as well as by the priors. In general, using standard forms for these distributions leads to a standard form for the posterior. We now demonstrate it for two cases of interest. (i) **Discrete to discrete**: Assume that node $u_i$ and its parents $\mathbf{pa}_i$ are discrete, so their connection is described by a probability table, $p(u_i = s \mid \mathbf{pa}_i = \mathbf{t}, \theta_i) = \theta_{i,s}^{\mathbf{t}} \geq 0$, where the parameters satisfy the normalization condition $\sum_s \theta_{i,s}^{\mathbf{t}} = 1$ for each $\mathbf{t}$. In this case, an appropriate prior on the parameters is a Dirichlet distribution with hyperparameters $\lambda_{i,s}^{\mathbf{t}}$: $p(\theta_i \mid m) = \prod_{\mathbf{t}} \mathcal{D}(\{\theta_{i,s}^{\mathbf{t}}\}; \{\lambda_{i,s}^{\mathbf{t}}\})$. To perform the average over $H$ in (9) we need the variational posterior over the hidden variables, which factorizes into $\prod_n q(u_i^n = s, \mathbf{pa}_i^n = \mathbf{t})$. It is straightforward to show that this averaging gives again a posterior which is a product of Dirichlet distributions with modified hyperparameters:

$$q(\theta_i \mid m) = \prod_{\mathbf{t}} \mathcal{D}(\{\theta_{i,s}^{\mathbf{t}}\}; \{\lambda_{i,s}^{\mathbf{t}} + N\bar{\pi}_{i,s}^{\mathbf{t}}\}) \,, \qquad (10)$$

where $\bar{\pi}_{i,s}^{\mathbf{t}} = \sum_n q(u_i^n = s, \mathbf{pa}_i^n = \mathbf{t} \mid m)/N$.

(ii) **Discrete to Normal**: Assume that node $u_i$ is continuous and Normally distributed conditioned on its parents $\mathbf{pa}_i$, which are discrete: $p(u_i = \mathbf{x} \mid \mathbf{pa}_i = \mathbf{t}, \theta_i) = \mathcal{N}(\mathbf{x}; \mu_{\mathbf{t}}, \Sigma_{\mathbf{t}})$, with the parameters $\theta_i = \{\mu_{\mathbf{t}}, \Sigma_{\mathbf{t}}\}$ having a Normal-Wishart prior $\mathcal{NW}(a_{\mathbf{t}}, \mathbf{B}_{\mathbf{t}}, \xi_{\mathbf{t}}, \beta_{\mathbf{t}})$ independently for each $\mathbf{t}$. It can be shown that the posterior $q(\theta_i \mid m)$ is also Normal-Wishart with modified hyperparameters:

$$q(\theta_i \mid m) = \prod_{\mathbf{t}} \mathcal{NW}(\mu_{\mathbf{t}}, \Sigma_{\mathbf{t}}; \qquad (11)$$
$$a_{\mathbf{t}} + N\bar{\pi}_{\mathbf{t}}, \mathbf{B}_{\mathbf{t}} + N\bar{\pi}_{\mathbf{t}}\mathbf{B}_{\mathbf{t}}', \xi_{\mathbf{t}}', \beta_{\mathbf{t}} + N\bar{\pi}_{\mathbf{t}}) \,,$$

where $\mathbf{B}_{\mathbf{t}}'$ and $\xi_{\mathbf{t}}'$ are determined from the first two moments of the hidden variable posterior $q(u_i^n = \mathbf{x} \mid \mathbf{pa}_i^n = \mathbf{t}, m)$, and $\bar{\pi}_{\mathbf{t}} = \sum_n q(\mathbf{pa}_i^n = \mathbf{t} \mid m)/N$.

Hence, in both cases the posterior has the same form as the prior. Notice that its covariance becomes $\mathcal{O}(1/N)$. In fact, (10,11) show that as the sample size $N$ increases, the influence of the prior on the form of the posterior diminishes. These results will be revisited below as specific models are being considered.

Next, to find the variational posterior over the hidden variables for a given structure $m$, we may try similarly to set $\delta \mathcal{F}/\delta q(H \mid m) = 0$ in (6), arriving at

$$\log q(H \mid m) = \langle \log p(Y, H \mid \Theta, m) \rangle_{\Theta \mid m} - \log z_m' \,, \quad (12)$$

where the average $\langle \cdot \rangle_{\Theta \mid m}$ is computed w.r.t. the parameter posterior $q(\Theta \mid m)$, and $z_m'$ is a (different) normalization constant. This procedure will be successful for some models, one of which is illustrated below. However, for many interesting models the resulting posterior will be quite difficult to work with, e.g., computing the normalization constant will be intractable, as well as performing the average in (8). In such cases we choose a parametric form for $q(H \mid \Phi, m)$ with a separate set of parameters $\Phi$, termed *variational parameters*, that are optimized to maximize $\mathcal{F}$. The crucial consideration in the choice of this posterior is that it allows performing all the required calculations analytically, while still providing a good approximation in the relevant sense. Below we shall demonstrate how this is done.

Finally, the posterior over the structures $m$ is similarly shown to be given by

$$\log q(m) = \langle \langle \log p(Y, H \mid \Theta, m) \rangle_{H \mid m}$$
$$+ \log \frac{p(\Theta \mid m)}{q(\Theta \mid m)} \rangle_{\Theta \mid m} - \log z \,. \quad (13)$$

As will be illustrated below, the parameter posteriors that emerge from VB turn out to have a parametric functional form, with these parameters (which should not be confused with the model parameters) being sufficient statistics (SS) computed from the data by an iterative, two-step, EM-like algorithm. In the E-step the hidden variable posterior is computed using the old SS; in the M-step the new SS are computed, updating the parameter posterior. in the large sample limit, this algorithm reduces to ordinary EM (Dempster et al. 1977).

### 2.5 Predictive Quantities and Labeling

The probability that a hypothesis is true given the data $D$ is determined by averaging over all models using their posteriors. For example, for density estimation applications, the predictive density for a new data vector $\mathbf{y}$ is

$$p(\mathbf{y} \mid Y) = \sum_{m \in M} \int d\Theta \sum_H$$
$$p(\mathbf{y} \mid H, \Theta, m) p(H, \Theta, m \mid Y) \,. \quad (14)$$

One approach is to directly replace the true posterior by the variational one $q(H, \Theta, m \mid Y)$. Unfortunately, since the variational posterior is designed to compute analytically averages over the logarithm of $p(\mathbf{y} \mid H, \Theta, m)$ rather than the actual distribution, additional approximations may be necessary.

However, the variational approach allows a rather attractive alternative route. Instead of considering the predictive density we consider its logarithm, given by $\log p(\mathbf{y} \mid Y) = \log p(Y') - \log p(Y)$, where $Y' = Y \cup \{\mathbf{y}\}$ is the augmented dataset. We now repeat the exact same steps used to compute the lower bound $\mathcal{F} \leq \log p(Y)$ (4,6), and compute $\mathcal{F}' \leq \log p(Y')$. This calculation requires only little additional effort, as the required posterior $q(H, \Theta, m \mid Y')$ is very close to the old one $q(H, \Theta, m \mid Y)$ which can be used for initialization. The predictive distribution is then given by

$$p(\mathbf{y} \mid Y) = e^{\mathcal{F}' - \mathcal{F}} . \quad (15)$$

In fact, in the large sample limit we obtain $\mathcal{F}' - \mathcal{F} = \sum_m q(m) \langle \langle \log p(\mathbf{y} \mid H, \Theta, m) \rangle_{H \mid m} \rangle_{\Theta \mid m}$ with no additional computation.

For other applications, such as unsupervised classification and blind source separation, the most likely value of a hidden variable $h_i$ given a new data vector $\mathbf{y}$ is required. This value is given by the MAP estimate $\hat{h}_i = \arg\max_{h_i} p(h_i \mid \mathbf{y}, Y)$. Again, one approach is to directly replace the true by the variational posterior, giving

$$\hat{h}_i = \arg\max_{h_i} \sum_{m \in M} q(m) \langle \langle p(\mathbf{y}, H \mid \theta, m) \rangle_{H^i \mid m} \rangle_{\Theta \mid m} , \quad (16)$$

where the average $\langle \cdot \rangle_{H^i \mid m}$ is performed w.r.t. $q(H \mid m)$ over all hidden variables after marginalizing it over $h_i$. Alternatively, we may compute the posterior for the augmented dataset as above, focus on the factor $q(H \mid m, Y')$ and marginalize over all hidden variables but $h_i$ to obtain $\hat{h}_i = \arg\max_{h_i} \sum_m q(h_i \mid m, Y')$.

A labeling problem may arise when computing a MAP estimate of hidden variables given new data. Consider two graph structures $m_1, m_2$ which contain the same hidden variables $h_i, h_j$, and assume both are invariant under the permutation $h_i \leftrightarrow h_j$. Then the node labeled $h_i$ in $m_1$ may be labeled $h_j$ in $m_2$, producing an incorrect estimate when summing over structures in (16). The same problem may arise from permutation of discrete values of a hidden variable (i.e., component label in mixture models). The honest way to avoid labeling problems is by incorporating appropriate prior information about the relevant hidden variables into the model to break its permutation invariance. A practical solution is to approximate the sum over all structures by a small number of most probable ones, and use post-processing to correct label switches. Of course, the problem is completely avoided if only the single most probable structure is used in place of the sum in (16).

## 3 Variational Bayes Mixture Models

### 3.1 Definitions

Mixture models constitute a useful tool for flexible density estimation. These models have been investigated and analyzed extensively (see, e.g., Titterington et al. 1985), and efficient methods exist for fitting a given model to data. However, the issue of determining the required number of mixture components is still an open problem. When viewed in the framework of unsupervised classification, this becomes the issue of determining the number of unobserved classes. While the Bayesian approach provides the solution in principle, no satisfactory practical algorithm has emerged from the application of involved sampling techniques (Richardson and Green 1997; Rasmussen 1999) and approximation methods (e.g., Cheeseman and Stutz 1995) to this problem. We now show that an elegant solution is provided by the VB approach.

We consider models of the form

$$p(\mathbf{y} \mid \Theta, m) = \sum_{s=1}^{m} p(\mathbf{y} \mid s, \Theta, m) p(s \mid \Theta, m) \quad (17)$$

(compare to (3)), where $m$ is the number of component, which determines the structure of the model. $\mathbf{y}$ denotes the observed data vector, $s = 1, ..., m$ is the hidden component label, $p(s \mid \Theta, m) = \pi_s$ the component probabilities which sum to one, and $p(\mathbf{y} \mid s, \Theta, m)$ the component distributions. Whereas our approach can be applied to arbitrary models, for simplicity we shall first consider the classical mixture model where the data are real-valued and the component distributions are Normal, $p(\mathbf{y} \mid s, \Theta, m) = \mathcal{N}(\mathbf{y}; \boldsymbol{\mu}_s, \boldsymbol{\Gamma}_s)$, where $\boldsymbol{\mu}_s$ is the mean and $\boldsymbol{\Gamma}_s$ the inverse covariance matrix. We use non-informative priors (this point will be revisited later) on the parameters $\Theta = \{\boldsymbol{\mu}_s, \boldsymbol{\Gamma}_s, \pi_s\}$, i.e., $p(\{\pi_s\})$ is flat, $p(\{\boldsymbol{\mu}_s\})$ is flat within an $m$-dim hypercube whose edge length $\to \infty$, and $p(\{\boldsymbol{\Gamma}_s\}) = \prod_s \mid \boldsymbol{\Gamma}_s \mid^{-1}$. Finally, we use a structure prior $p(m) = 1/K$ for $1 \leq m \leq K$ with $K$ being the maximal number of components.

### 3.2 Learning Algorithm

We can now follow the steps outlined above for a dataset $Y = \mathbf{y}_{1:N}$ and derive the variational posterior distributions over parameters $q(\Theta \mid m)$, component label $q(s_{1:N} \mid m)$, and structure $q(m)$. When doing this, we find that the parameter posterior factorizes into $q(\Theta \mid m) = \prod_s q(\boldsymbol{\mu}_s, \boldsymbol{\Gamma}_s \mid m) q(\{\pi_s\} \mid m)$, as predicted from (9). The mean and inverse covariance are jointly Normal-Wishart; note that they factorize over $s$ as well. The component probabilities are jointly Dirichlet. These results are consistent with the general properties (10,11). To make the results more transparent we further restricted the parameter posterior to factorize the mean from the covariance (although all calculations can be fully carried out using their joint distribution), arriving at

$$\begin{aligned} q(\boldsymbol{\mu}_s \mid m) &= \mathcal{N}(\boldsymbol{\mu}_s; \bar{\boldsymbol{\mu}}_s, N\bar{\pi}_s \bar{\boldsymbol{\Gamma}}_s) \,, \\ q(\boldsymbol{\Gamma}_s \mid m) &= \mathcal{W}(\boldsymbol{\Gamma}_s; a_s, \mathbf{B}_s) \,, \\ q(\{\pi_s\} \mid m) &= \mathcal{D}(\{\pi_s\}; N\bar{\pi}_1 + 1, ..., N\bar{\pi}_m + 1) \,, \end{aligned} \quad (18)$$

where $a_s = N\bar{\pi}_s/2$ and $\mathbf{B}_s = a_s \bar{\boldsymbol{\Gamma}}_s^{-1}$. Hence the means are Normal, the inverse covariances are Wishart, and the component probabilities remain Dirichlet. The parameters appearing in (18) will be defined shortly.

The label posterior factorizes over the data instances, $q(s_{1:N} \mid Y, m) = \prod_n q(s_n \mid \mathbf{y}_n, m)$ for obvious reasons, and is given for instance $n$ by

$$\begin{aligned} q(s_n \mid m) &= \frac{1}{z_n} \tilde{\pi}_s \times \mid \frac{\tilde{\boldsymbol{\Gamma}}_s}{2\pi} \mid e^{-(\mathbf{y}_n - \bar{\boldsymbol{\mu}}_s)^T \tilde{\boldsymbol{\Gamma}}_s (\mathbf{y}_n - \bar{\boldsymbol{\mu}}_s)/2)} \\ &\quad \times e^{-d/2N\bar{\pi}_s} \,, \end{aligned} \quad (19)$$

where $d$ is the data dimensionality, $z_n$ a normalization constant, $\tilde{\pi}_s = \exp(\Psi(N\bar{\pi}_s + 1) - \Psi(N + m))$, $\tilde{\boldsymbol{\Gamma}}_s = \mid \mathbf{B}_s \mid^{-1} \exp(d\Psi(a_s))$, and $\Psi$ is the psi (digamma) function.

The VB approach has therefore led us to an EM-like algorithm for each structure $m$, where in the E-step we learn the label posterior (19), and in the M-step we learn the parameter posteriors (18). In fact, we obtain the following learning algorithm for the parameters, which determine the sufficient statistics (SS) of these distributions. First we initialize $\bar{\pi}_s, \bar{\boldsymbol{\mu}}_s, \bar{\boldsymbol{\Gamma}}_s$ to appropriate values. Next we compute $q(s_n \mid m)$ from (19). Then we compute the new SS by

$$\bar{\pi}_s = \langle 1 \rangle_{s \mid m} \,, \quad \bar{\boldsymbol{\mu}}_s = \frac{1}{\bar{\pi}_s} \langle \mathbf{y} \rangle_{s \mid m} \,, \quad (20)$$

$$\bar{\boldsymbol{\Gamma}}_s = (1 - \frac{1}{N\bar{\pi}_s}) \left( \frac{1}{\bar{\pi}_s} \langle (\mathbf{y} - \bar{\boldsymbol{\mu}}_s)(\mathbf{y} - \bar{\boldsymbol{\mu}}_s)^T \rangle_{s \mid m} \right)^{-1} \,,$$

where $\langle \cdot \rangle_{s \mid m}$ implies averaging w.r.t. the label posterior, i.e., $\langle f(\mathbf{y}) \rangle_{s \mid m} = \sum_n f(\mathbf{y}_n) q(s_n \mid m)/N$. These steps are repeated until convergence; the case $\bar{\pi}_s \leq 1/N$ will be commented upon below.

Several remarks deserve to be made at this point. (a) Whereas no prior assumptions about them have been made, the parameters posteriors that emerge from applying VB to the mixture model using uninformative priors have non-trivial functional forms that are intuitively appealing. More importantly, in general they are not Normal, unlike those resulting from the Laplace approximation. (b) In fact, no complications would appear if we generalize the priors to have the same functional form of the corresponding posteriors (besides the appearance of appropriate hyperparameters). (c) The quantities $\bar{\pi}_s, \bar{\boldsymbol{\mu}}_s, \bar{\boldsymbol{\Gamma}}_s$ learned by the rules (20) are the means of the posterior distributions (18) over the parameters (more precisely, $\bar{\pi}_s$ are the mean component probabilities only in the large sample limit). The covariances of these posteriors are $\mathcal{O}(1/N)$. (d) In the large sample limit, the posteriors collapse onto their means, and also $\tilde{\pi}_s = \bar{\pi}_s$, $\tilde{\boldsymbol{\Gamma}}_s = \bar{\boldsymbol{\Gamma}}_s$. Therefore in this limit we recover the ordinary EM algorithm. (e) Most strikingly, when the number of data vectors assigned to component $s$ is one or less, i.e., $\bar{\pi}_s \leq 1/N$, it can be shown that the rules (19–20) are replaced by $q(s_n = s) = \bar{\pi}_s = 0$, effectively declaring component $s$ nonexistent. This property is important since it protects our algorithm from the following well-known problem with the ordinary EM algorithm for mixture models: There, one component may become centered at a single data vector, sending its covariance to zero and the model likelihood to infinity; the resulting wrong model is thus assigned a higher likelihood than the correct one. The VB algorithm automatically eliminates such a component.

Finally, once the posterior distributions over the parameters and label conditioned on structure have been obtained, the posterior distribution over model structure $q(m)$ is given by

$$q(m) = \frac{1}{z} \exp \left( \langle \log \frac{p(\mathbf{y}_{1:N}, s_{1:N}, \Theta \mid m)}{q(s_{1:n}, \Theta \mid m)} \rangle_{s, \Theta \mid m} \right) \,, \quad (21)$$

where $\langle \cdot \rangle_{s, \Theta \mid m}$ refers to averaging w.r.t. $q(s_{1:n}, \Theta \mid m)$. The resulting closed-form expression is omitted.

Also omitted is the expression we obtained for the predictive density (15); we point out, however, that it is not a mixture of Normal distributions.

### 3.3 Results

I applied the VB mixture model algorithm to several toy problems; Fig. 1 presents the results on two of them. In the first (top) 600 data points were generated from a 2-dim mixture model with Normal components, whose covariances are represented by the ellipses on the top left panel. The algorithm was then applied with maximum number of components $K = 10$; the resulting log-posterior over the number of components is shown on the top right panel, indicating that the posterior is sharply peaked at the correct value $m = 3$.

In the second problem (bottom), the VB algorithm was applied to 800 data points generated from a 3-dim noisy spiral. Here there is no 'correct' number; the resulting posterior (whose logarithm is plotted on the bottom right panel) is peaked at $m = 11$. The means of the resulting posterior over the covariances for this case are represented by their axes (bottom left). Larger numbers of components were observed to produce overlaps. The VB mixture is currently being applied to the task of handwritten character recognition.

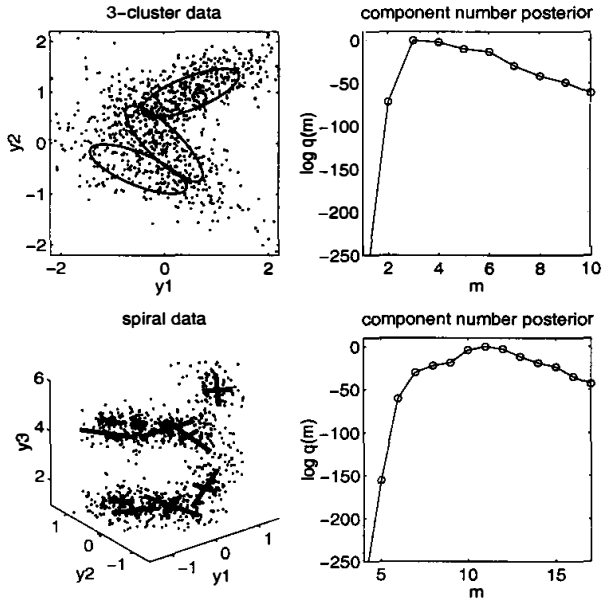

Figure 1: Application of the VB mixture model to density estimation. Top: Data generated from a 3-component model (left) and the resulting log-posterior over the number of components. Bottom: Data generated from a noisy spiral and the means of the covariance posterior corresponding to an 11-component model (left), and the log-posterior over the number of components.

## 4  Blind Source Separation

### 4.1  Definitions

In the blind source separation (BSS) problem, a.k.a. independent component analysis (ICA) (Jutten and Herault 1991; Bell and Sejnowski 1995; Cardoso and Lahed 1996; Attias and Schreiner 1998), one is presented with multivariable time series data. It is assumed that these data, which are generally correlated, arise from several individual source signals that are mutually statistically independent. The sources are unobservable and are mixed together by an unknown linear transformation, corrupted by unobservable noise. The task is to recover the sources from the data. A successful solution to this problem will have many applications in areas involving processing of multisensor signals, such as speech recognition and enhancement, analysis and classification of biomedical data, target localization and tracking by radar and sonar devices, and wireless communication.

Let $x_j^n$ denote the signal emitted by source $j = 1 : m$ at time $n$, and let $y_i^n$ denote the signal received at sensor $i = 1 : d$ at the same time. In the instantaneous mixing version of this problem, we assume that the two are linearly related: $\mathbf{y}_n = \mathbf{A}\mathbf{x}_n + \mathbf{u}_n$, where the $d \times m$ matrix $\mathbf{A}$ is termed the *mixing matrix*, and $u_i^n$ are zero-mean Gaussian noise signals with inverse variance $\lambda_i$. We also assume that we have good approximations for the independent source densities $p_i(x_i)$. We shall use the model $p_i(x_i) = \cosh^{-2}(x_i/2)/2$, which has been shown to be accurate for the purpose of separating speech sources (Bell and Sejnowski 1995; Attias and Schreiner 1998). Thus, the graph we consider is given by

$$p(\mathbf{x} \mid m) = \prod_{j=1}^{m} \frac{1}{2} \cosh^{-2}(\frac{x_j}{2}) ,$$

$$p(\mathbf{y} \mid \mathbf{x}, \mathbf{A}, m) = \prod_{i=1}^{d} \mathcal{N}(y_i; \sum_{j=1}^{m} A_{ij} x_j, \lambda_i) . \quad (22)$$

In terms of (3), we have hidden variables $H = \{x_j\}$, parameters $\Theta = \{A_{ij}\}$, and structure $m$ determined by the number of sources.

Most existing ICA algorithms address the simplified case where the noise vanishes and the mixing matrix is square invertible, so the number of sensors equals the number of sources. Furthermore, $m$ is assumed known in advance. Using maximum likelihood, one computes the distribution assigned to the data by the model $(\mathbf{A}, m)$, $p(\mathbf{y} \mid \mathbf{A}, m) = \mid \mathbf{A} \mid^{-1} p(\mathbf{x} \mid m)$, and chooses $\mathbf{A}$ to maximize it. The sources are then recovered via $\mathbf{x} = \mathbf{A}^{-1}\mathbf{y}$.

The more general case of non-square mixing and non-zero noise is harder, since one has to compute $p(\mathbf{y} \mid \mathbf{A}, m) = \int d\mathbf{x}\, p(\mathbf{y} \mid \mathbf{x}, \mathbf{A}, m) p(\mathbf{x} \mid m)$, where the $m$-dim integration is non-trivial due to the non-Gaussian nature of the sources. Lewicki and Sejnowski (1998) integrated over the sources using the Laplace approximation. Attias (1999a) solved this problem by modeling each source density by a 1-dim mixture of Gaussians, which allows the above integral to be calculated analytically. The sources are then reconstructed by a MAP estimate: $\hat{\mathbf{x}} = \arg\max_{\mathbf{x}} p(\mathbf{x} \mid \mathbf{y}, \mathbf{A}, m)$. This approach results in an EM algorithm that learns both the mixing and noise covariance matrices, as well as the source distributions, from noisy data. Since the computational complexity of the algorithm increases exponentially with the number of sources, the large $m$ case is treated in (Attias 1999a) by a structured variational approximation (Ghahramani and Jordan 1997).

However, in realistic cases the observed data is generated by an unknown number of sources $m$. Here we exploit the VB approach to compute the posterior distribution over $m$ from a dataset $Y$ of sensor signals.

We point out that realistic situations include many additional complications, such as multipath propagation and reverberant conditions (see (Attias and Schreiner 1998) for a treatment of the zero-noise convolutive blind separation problem), as well as non-stationarity; these issues are beyond the scope of the present paper.

### 4.2 Learning Algorithm

For the prior distribution on the mixing matrix, we choose the elements $A_{ij}$ to be independent, zero-mean Normal variables with precision $\alpha$ as a single hyperparameter, i.e.,

$$p(\mathbf{A} \mid \alpha, m) = \left(\frac{\alpha}{2\pi}\right)^{dm/2} \exp\left(-\frac{\alpha}{2} \sum_{ij} H_{ij}^2\right) . \quad (23)$$

This prior becomes uninformative in the limit $\alpha \to 0$. For simplicity, we keep $\alpha$ and the noise precisions $\lambda_i$ as hyperparameters, although VB can treat them in a full probabilistic manner with little added effort. The structure prior employed is $p(m) = 1/K$ for $1 \le m \le K$ with $K$ being the maximal number of sources.

Following the discussion of section 2.4, we find that the mixing matrix posterior is Normal,

$$q(\mathbf{A} \mid m) = \mathcal{N}(\mathbf{A}; \bar{\mathbf{A}}, \mathbf{\Sigma}) , \quad (24)$$

whose mean and covariance are given by

$$\begin{aligned}
\bar{A}_{ij} &= \left[\mathbf{C}_{yx}(\mathbf{C}_{xx}^i)^{-1}\right]_{ij} , \\
\Sigma_{ij,kl} &= \frac{1}{\lambda_i N} \left(\mathbf{C}_{xx}^i\right)^{-1}_{jl} \delta_{ik} , \\
\mathbf{C}_{xx}^i &\equiv \mathbf{C}_{xx} + \frac{\alpha}{\lambda_i N}\mathbf{I} ,
\end{aligned} \quad (25)$$

where that $\Sigma_{ij,kl}$ is the expectation of $(A_{ij} - \bar{A}_{ij})(A_{kl} - \bar{A}_{kl})$. Viewing $A_{ij}$ as a $dm \times 1$ vector formed by concatenating the columns of $\mathbf{A}$ into a large column, note that $\mathbf{\Sigma}$ has a block-diagonal form consisting of $d$ blocks of dimension $m \times m$. The correlation matrices are $\mathbf{C}_{yx} = \langle \mathbf{y}\mathbf{x}^T \rangle_{x|m} = \sum_n \mathbf{y}_n \rho_n^T / N$ and $\mathbf{C}_{xx} = \langle \mathbf{x}\mathbf{x}^T \rangle_{x|m} = \sum_n (\rho_n \rho_n^T + \Gamma^{-1})/N$; the averages are computed w.r.t. the source posterior $q(\mathbf{x} \mid m)$ (26). We point out that in the large sample limit, the covariance of $\mathbf{A}$ vanishes and its mean becomes $\bar{\mathbf{A}} = \mathbf{C}_{yx}(\mathbf{C}_{xx})^{-1}$, a form appearing in the ordinary EM algorithms for factor analysis (Rubin and Thayer 1982) and independent factor analysis (Attias 1999a).

However, the source posterior cannot be obtained by directly optimizing $\mathcal{F}$ (see (12)), due to the non-Gaussian nature of the sources. Instead we use two variational tricks. First, noting that the source posterior factorizes over instances, i.e., $q(\mathbf{x}_{1:N} \mid m, Y) = \prod_n q(\mathbf{x}_n \mid m, \mathbf{y}_n)$, we choose a Normal distribution at each instance $n$,

$$q(\mathbf{x}_n \mid m) = \mathcal{N}(\mathbf{x}_n; \rho_n, \Gamma_n) , \quad (26)$$

where the mean $\rho$ and (general) inverse covariance $\Gamma$, termed *variational parameters*, may depend on the data $\mathbf{y}_n$, and will be adapted to help this posterior best approximate the optimal one. Second, in order to adapt them we must compute the expected value of $\log p(Y, \mathbf{x} \mid \mathbf{A}, m)$ under this posterior, which poses a difficulty, again due to the form of $p(\mathbf{x} \mid m)$. To overcome it, we exploit Jensen's inequality to compute a bound on this quantity:

$$\begin{aligned}
\langle \log p(\mathbf{x} \mid m) \rangle_{x|m} &= -2 \sum_{i=1}^{m} \langle \log \frac{\cosh(x_i/2)}{2} \rangle_{x|m} \\
&\ge -\frac{1}{4}\text{Tr}\Gamma^{-1} - 2 \sum_{i=1}^{L} \log \frac{\cosh(\rho_i/2)}{2} .
\end{aligned} \quad (27)$$

In general, the accuracy of this lower bound depends on the variational parameters $\rho$ and $\Gamma$, especially on the latter: Note that in the zero-noise case ($\lambda_i \to \infty$), $\Gamma^{-1}$ vanishes and the bound is exact. I found experimentally that, for the distributions of $\rho$ and $\Gamma$ arising in the cases treated in the present paper, the mean error of the bound is smaller than 4%.

Given the posterior over $\mathbf{A}$, we can now set $\partial \mathcal{F}/\partial \rho_n = 0$ and derive the fixed point equation

$$\bar{\mathbf{A}}^T \mathbf{\Lambda}(\mathbf{y}_n - \bar{\mathbf{A}}\rho_n) - \tanh \frac{\rho_n}{2} = \frac{\sum_i (\mathbf{C}_{xx}^i)^{-1}}{N} \rho_n , \quad (28)$$

where $\mathbf{\Lambda} = \text{diag}(\lambda_1, ..., \lambda_d)$. This equation can be solved iteratively for each $n$, using the initial value $\rho_n = (\mathbf{A}^T \mathbf{A})^{-1} \mathbf{A} \mathbf{y}_n$ which is exact in the limit of low noise and large sample size. The variational precision matrix turns out from $\partial \mathcal{F}/\partial \Gamma_n = 0$ to be $n$-independent:

$$\Gamma_n = \Gamma \equiv \bar{\mathbf{A}}^T \mathbf{\Lambda} \bar{\mathbf{A}} + \frac{1}{2}\mathbf{I} + \frac{\sum_i (\mathbf{C}_{xx}^i)^{-1}}{N} . \quad (29)$$

Finally, optimizing the hyperparameter $\alpha$ gives

$$\alpha = \frac{1}{dm}\text{Tr}\left(\bar{\mathbf{A}}^T \bar{\mathbf{A}} + \sum_i \frac{(\mathbf{C}_{xx}^i)^{-1}}{\lambda_i N}\right) . \quad (30)$$

An optimization rule for the hyperparameters $\lambda_i$ can similarly be derived but is omitted. We also omit the structure posterior $q(m)$.

Hence, like in the mixture model case, VB led to an EM-like algorithm for each $m$, where the E-step learns the source posterior (26), and the M-step learns the parameter posterior (24). The algorithm actually learns the SS of these distributions as follows. First we initialize $\bar{\mathbf{A}}$, $\boldsymbol{\Sigma}$, $\lambda_i$, $\alpha$ to appropriate values. Next we compute the SS of $q(\mathbf{x} \mid m)$ using (28–29). Then we compute the new SS of $q(\mathbf{A} \mid m)$ from (25,30). These steps are repeated until convergence.

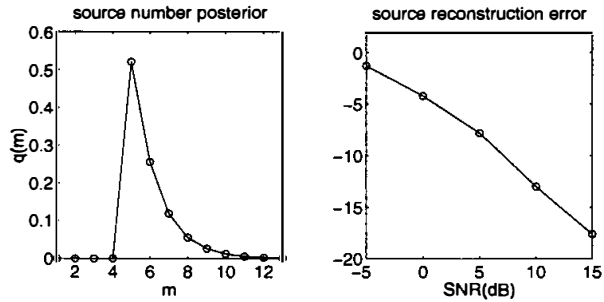

Figure 2: Left: Application of the VB source separation algorithm to 11-dim data generated by linear mixing of 5 speech and music signals. Left: The resulting posterior over the number of sources. Right: log-error in reconstructed sources w.r.t. the original ones for different noise levels.

We remark that a Variational Bayes algorithm for the more conventional method of factor analysis can straightforwardly be derived. In that case, the sources are Normal and the source posterior (26) is actually optimal (within the VB framework). The resulting equation for $\rho$ can be solved in a single iteration. However, as is well known, the Gaussian nature of the sources prevents factor analysis from performing source separation.

### 4.3 Results

I applied the VB source separation algorithm to 11-dim data generated by mixing 5 speech and music signals obtained from commercial CDs. Each signal was 1sec long at sampling frequency 8.82kHz. The signals were mixed by a random 11 × 5 mixing matrix, and different levels of Gaussian noise were added. The posterior over the number of sources found by the algorithm is plotted in Fig. 2 (left), and is peaked at the correct value of $m = 5$. The sources were then reconstructed from the data using a MAP estimate. The log-error of the reconstructed w.r.t. the original sources is plotted for different signal-to-noise (SNR) levels in Fig. 2 (right), and is seen to decrease with increasing SNR as expected. Additional experiments with different numbers of sources and of sensors gave similar results.

## 5 Hierarchical Mixtures and Probable Structures

Whereas the integration over all model parameters and structures is tractable in the two models discussed above, in more complicated models such a full Bayesian treatment is practically impossible. Consider the hierarchical mixture model constructed as follows. Each mixture component $s = 1, ..., l$ has a probability $p(s \mid l) = \pi_s$, and a distribution of a blind separation model with $k_s$ sources and a $d \times k_s$ mixing matrix $\mathbf{A}_s$. Hence, the graphical model is described by the joint distribution

$$\begin{aligned} p &= p(\mathbf{y} \mid \mathbf{x}, \mathbf{A}, \pi_{1:l}, s, k_{1:l}, l) p(\mathbf{x} \mid s, k_s) p(s \mid l) \\ &\quad p(\mathbf{A} \mid k_s, s) p(\pi_{1:l} \mid l) p(k_{1:l} \mid l) p(l) \,. \end{aligned} \quad (31)$$

This model is potentially useful for pattern recognition on speech and image data. The reason is that these data typically have long-tailed distributions, which are modeled more efficiently by exponential rather than Normal component distributions. However, denoting by $K$ the maximal number of sources for each component and by $L$ the maximal number of components, we have $K^L$ possible structures for this model.

A simple way to obtain a polynomial time algorithm is to include only the most probable structure and possibly a few neighboring structures. Formally, this procedure amounts to making the *factorized* variational approximation

$$q(k_{1:l}, l) = \delta_{l,l^0} \prod_{s=1}^{l} \left( \sum_r w_s^r \delta_{k_s, k_s^0 + r} \right) , \quad (32)$$

with $r$ covering a small range of $r_0$ numbers including zero and $w_s^r \geq 0$ satisfying $\sum_r w_s^r = 1$. The form (32) allows only a single number of components $l^0$, and restricts each component $s$ to a range of $r_0$ possible source numbers about $k_s^0$ with probabilities $w_s^r$. The quantities $l^0$, $k_{1:l}^0$, $w_{1:l}^r$ are variational parameters which depend on the dataset $Y$; their optimization amounts to performing a local search in structure space for the most probable structures (although the $w_s^r$ may be fixed). Of course, alternative variational structure posteriors are possible.

## 6 Conclusion

This paper developed an approximation scheme for Bayesian inference in graphical models with hidden variables, and demonstrated it on density estimation and blind source separation tasks. A comparison of the accuracy of VB with that of the Laplace approximation against a Monte Carlo standard would be an important undertaking. It will be exciting to apply the VB framework to complex Bayesian networks (e.g., Attias 1999b), including dynamic models, and demonstrate its performance on real-world tasks such as speech recognition and scene analysis.